\documentclass[11pt]{article}

\usepackage{acl}
\usepackage{hyperref}
\usepackage[hyphenbreaks]{breakurl}
\usepackage{ifthen}
\newboolean{long}
\setboolean{long}{true}%
\usepackage{url}

\usepackage{times}
\usepackage{latexsym}

\usepackage{adjustbox}
\usepackage[table,dvipsnames]{xcolor}
\usepackage{tikz}
\usepackage{caption}
\usepackage{subcaption}
\usepackage[]{microtype}

\usepackage{xstring}
\usepackage{todonotes}
\usepackage{url}
\usepackage{graphicx}
\graphicspath{{fig}}
\usepackage{pgfplots}
\pgfplotsset{width=10cm,compat=1.9}
\usepackage{lingmacros}
\usepackage{lipsum}

\usepackage{amsmath,amsfonts,amssymb}

\usepackage{collcell}
\usepackage{booktabs}
\usepackage{array}
\usepackage{multicol}
\usepackage{multirow}
\usepackage{enumitem}

\usepackage{xspace}
\newcommand{\bloom}[0]{\textsc{Bloom}\xspace}
\newcommand{\opt}[1]{\textsc{OPT#1}\xspace}
\newcommand{\gpttwo}[0]{\textsc{GPT-2}\xspace}
\newcommand{\gptthree}[0]{\textsc{GPT-3}\xspace}

\newcommand{\palm}[0]{\textsc{Palm}\xspace}

\newcommand{\chatgpt}[0]{\textsc{ChatGPT}\xspace}
\newcommand{\xglm}[0]{\textsc{XGLM}\xspace}
\newcommand{\glm}[0]{\textsc{GLM}\xspace}
\newcommand{\alexatm}[0]{\textsc{AlexaTM}\xspace}
\newcommand{\tfive}[0]{\textsc{T5}\xspace}

\newcommand{\mtzero}[0]{\textsc{mT0-xxl}\xspace{}}
\newcommand{\tzero}[0]{\textsc{T0}\xspace{}}

\newcolumntype{R}[2]{%
    >{\adjustbox{angle=#1,lap=\width-(#2)}\bgroup}%
    l%
    <{\egroup}%
}

\newcommand*{\MinNumber}{0}
\newcommand*{\MidNumber}{5}
\newcommand*{\MaxNumber}{150}

\newcommand{\ApplyGradient}[1]{%
    \IfDecimal{#1}{%
      \edef\mynum{#1}%
      \ifdim #1 pt > \MaxNumber pt\relax
        \edef\mynum{\MaxNumber}%
      \else
        \ifdim #1 pt < \MinNumber pt\relax
          \edef\mynum{\MinNumber}%
        \fi
      \fi
      \ifdim \mynum pt > \MidNumber pt
        \pgfmathsetmacro{\PercentColor}{max(min(100.0*(\mynum - \MidNumber)/(\MaxNumber-\MidNumber),100.0),0.00)}%
        \xdef\PercentColorr{\PercentColor}%
        \cellcolor{MidnightBlue!\PercentColorr!Yellow}#1%
      \else
        \pgfmathsetmacro{\PercentColor}{max(min(100.0*(\MidNumber - \mynum)/(\MidNumber-\MinNumber),100.0),0.00)}%
        \xdef\PercentColorr{\PercentColor}%
        \cellcolor{SpringGreen!\PercentColorr!Dandelion}#1%
      \fi	%
    }{\textbf{#1}}%
}
\newcolumntype{G}{>{\collectcell\ApplyGradient}c<{\endcollectcell}}

\title{Investigating the Translation Performance of a Large Multilingual Language Model: the Case of \bloom{} }

\ifthenelse{\boolean{long}}{
\author{Rachel Bawden\\
  Inria, Paris, France\\
  \\
  {\tt rachel.bawden@inria.fr}  \And
  François Yvon\\
  Université Paris-Saclay, CNRS, LISN \\
  \\
  {\tt francois.yvon@cnrs.fr}}
}{
\author{Rachel Bawden\\
  Inria, Paris, France\\
  \\
  {\tt rachel.bawden@inria.fr}  \And
  François Yvon\\
  Université Paris-Saclay, CNRS, LISN \\
  \\
  {\tt francois.yvon@cnrs.fr}}
}
\date{}

\begin{document}
\maketitle
\begin{abstract}
The NLP community recently saw the release of a new large open-access multilingual language model, \bloom \citep{lescao-etal-2022-bloom} covering 46 languages. We focus on \bloom's multilingual ability by evaluating its machine translation performance across several datasets (WMT, Flores-101 and DiaBLa) and language pairs (high- and low-resourced). Our results show that 0-shot performance suffers from overgeneration and generating in the wrong language, but this is greatly improved in the few-shot setting, with very good results for a number of language pairs.  We study several aspects including prompt design, model sizes, cross-lingual transfer and the use of discursive context.
\end{abstract}

\maketitle

\section{Introduction}\label{sec:intro}
Large language models (LLMs) trained at scale with simple objectives have been found to achieve results that match  dedicated systems on numerous NLP tasks \citep{gpt2}, as long as tasks are formulated as text generation though ``prompting'' %
\citep{liu-etal-2023-pretrain}. %
LLMs' multi-task performance can even be improved with ``instruction'' fine-tuning \citep{sanh-etal-2022-multitask,muenninghoff-etal-2022-bloomz}, few-shot priming, and better strategies to select or learn prompts \citep{petroni-etal-2019-language,shin-etal-2020-autoprompt,schick-schutze-2021-just,lester-etal-2021-power,wei-etal-2022-flan}. In multilingual settings, their performance on machine translation (MT) tasks, as measured by automatic scores, is often close to state of the art, even when mostly trained on monolingual data \citep{gpt3}. Moreover, prompting-based MT offers the prospect of better control of outputs, e.g.\ in terms of quality, style and dialect \citep{garcia-firat-2022-using}. However, these abilities remain poorly understood, as LLM analyses primarily focus on their multitask rather than multilingual ability (see however \citep{vilar-etal-2022-prompting,zhang-etal-2023-prompting,moslem-etal-2023-adaptive}, which we discuss in Section~\ref{sec:related}).

In this work, we focus on the MT performance of \bloom \citep{lescao-etal-2022-bloom}, a (family of) open-access multilingual LLM(s), designed and trained by the collaborative BigScience project.%
\footnote{\url{https://hf.co/bigscience/bloom}} Our main aims are to (i)~evaluate \bloom's zero- and multi-shot behaviour, (ii)~study the effect of prompt design, %
(iii)~evaluate a diverse set of language pairs %
and (iv)~assess its ability to use linguistic context. Our main conclusions, which extend those  in \citep{lescao-etal-2022-bloom}, are (i)~0-shot ability is blighted by overgeneration and generating in the wrong language, (ii)~using few-shot improves both issues, with results much closer to state of the art across datasets and language pairs, (iii)~there are clear transfer effects, with high scores for languages not officially seen in training, and successful transfer across language pairs via few-shot examples and (iv)~although linguistic context does not lead to higher scores, there is evidence that \bloom's translations are influenced by it. %
We release our code and translation outputs.\ifthenelse{\boolean{long}}{\footnote{\url{https://github.com/rbawden/mt-bigscience}}}{\footnote{\url{https://github.com/rbawden/mt-bigscience}}}

\section{Related work}\label{sec:related}
\ifthenelse{\boolean{long}}{%
Since the early attempts to use language models (LMs) as multi-task learners \citep{mccann-etal-2018-decathlon}, MT has been a task of choice to gauge LMs' multilingual ability. Results for the zero- and few-shot ability of LMs were discussed for both \gpttwo and \gptthree \citep{gpt2,gpt3}, which is especially intriguing as they were trained primarily on monolingual (English) data. These results have since been confirmed for other monolingual LMs such as \tfive \citep{raffel-etal-2020-exploring} and multilingual LMs such as \xglm \citep{lin2021xglm}, \palm \citep{chowdhery-etal-2022-palm}, and \alexatm \citep{soltan-etal-2022-alexatm}. However, the focus has mainly been on global multi-task performance;  often only a small part of the discussion is devoted to MT. Moreover, results are often only reported for a few well-resourced language pairs (e.g.\ English-French and English-German), and the scores reported (mostly BLEU), are hard to compare due to a non-systematic use of standardised evaluation protocols and metrics.\footnote{See the discussion at \url{http://blog.benjaminmarie.com/2/comparing-uncomparable.html} of these differences, and an attempt to reconstruct consistent scores.}

There are however some in-depth analyses of MT performance of LLMs, each focusing on a specific LM's performance in a true multilingual setting with respect to prompt design and number of few-shots.
For instance, \citet{vilar-etal-2022-prompting} reevaluate the MT performance of the multilingual \palm \citep{chowdhery-etal-2022-palm}, focusing notably on the selection of few-shot examples. 
Consistent with our findings, they determine that prompt choice becomes unimportant in few-shot settings and that using few-shot examples increases performance with diminishing returns for $k>5$ examples, using BLEURT and BLEU scores, as well as the results of a human evaluation. 
They find that the quality of few-shot examples has a large impact on performance. However, even with good prompts, \palm lags a couple of points behind state-of-the-art MT systems, especially when translating from English, notable due to adequacy problems. %
\citet{zhang-etal-2023-prompting} focus on the evaluation of \glm-130B, a bilingual (Chinese and English) LLM \citep{glm}. %
Their main conclusions %
are also consistent with ours: (a)~zero-shot performance varies greatly across different prompts, (b)~increasing the number of prompts from 0 to 20 yields consistent improvements in performance, again with variance across instructions, and (c)~finding the best few-shot example selection policy is difficult. It seems that having good and long examples, for instance, may help, even though none of the criteria explored in this study seem to provide any systematic improvement. A last point worth mentioning is that prompting with monolingual data hurts performance, but that using pseudo-parallel data obtained with back-translation \citep{bojar-tamchyna-2011-improving} is an effective workaround.

\citet{moslem-etal-2023-adaptive} evaluate OpenAI's \gptthree \citep{gpt3}\footnote{Version: \texttt{text-davinci-003} model.} with sampling-based decoding and a prompt resembling our own \texttt{xglm-source+target} prompt. They report strong zero-shot behaviour using multiple metrics, plus clear improvements with an increased number of shots for the well-resourced languages, less so for the only low-resource language in their lot (Kinyarwanda). The main novelty of this study is to use prompting as a vehicle to perform local adaptation and to ensure terminological consistency. For this, they use fuzzy matches from a translation memory as well as MT outputs to build their prompts, yielding results that both outperform their zero-shot system, but also their initial MT engine. Additionally inserting terms and their translation in the instruction yields supplementary improvements.

Finally note the preliminary evaluation of \chatgpt in \citep{jiao-etal-2023-is-chatgpt}, which reports interesting insights regarding the multilingual abilities of this model, as well as proposing innovative techniques to generate (artificial) prompts and to use pivoting in prompting. Similar to ours, this study considers multiple test domains such as news (WMT) and Wikipedia (Flores). A more in-depth analysis of the same model can be found in \citep{hendy-etal-2023-howgood}, which confirms \chatgpt's strong translation abilities, at least for ``well-resourced''\footnote{A rather slippery concept in this context, as the content of the training data is not fully known and seems to mostly comprise English texts.} language pairs. Document-level scores are also reported, as well as human evaluations and qualitative analyses.}%
{%
  Since the early attempts at using language models (LMs) as multi-task learners \citep{mccann-etal-2018-decathlon}, MT has been a task of choice to gauge LMs' multilingual ability. Results for the zero- and few-shot ability of LMs were discussed for both \gpttwo and \gptthree \citep{gpt2,gpt3}. These results have since been confirmed for other monolingual LMs such as \tfive \citep{raffel-etal-2020-exploring} and multilingual LMs such as \xglm \citep{lin2021xglm}, \palm \citep{chowdhery-etal-2022-palm}, and \alexatm \citep{soltan-etal-2022-alexatm}. However, the focus of these studies has mainly been multi-task performance, with  little analysis of MT results. Moreover, results are often only for a few well-resourced language pairs (e.g.\ English-French and English-German) and the scores reported (mostly BLEU) not always easy to compare.

  There are however a number of recent in-depth analyses of MT performance of LLMs, each focusing, like we do, on one specific LM. Most discuss, as we do, the variation of performance with respect to prompt design and number of few-shots examples. This is the case for example of \citet{chowdhery-etal-2022-palm}, who reanalyse \palm's translations and \citet{zhang-etal-2023-prompting}, who focus on \glm-130B, a bilingual (Chinese and English) LLM \citep{glm}. Consistent with our findings, these studies observe commandable zero-shot performance, with a great variation depending on prompt choices, which tends to diminish when more prompts are used. Using more than 5-10 examples, however, seems to bring very little return. The choice of few-shot examples does make a difference, as also observed by \citet{moslem-etal-2023-adaptive} in their evaluation of OpenAI's \gptthree \citep{gpt3}.\footnote{Version: \texttt{text-davinci-003} model.} The study considers a single prompt resembling our \texttt{xglm-source+target} prompt, but varies the strategy used to select examples, showing that prompting can effectively serve as a vehicle to perform local adaptation and to enforce terminological consistency. Finally it is worth mentioning the preliminary evaluation of \chatgpt in \citep{jiao-etal-2023-is-chatgpt}, and the more detailed one in \citep{hendy-etal-2023-howgood}, which confirms the strong translation abilities of this model, at least for ``well-resourced''\footnote{A rather slippery concept in this context as the training data content, seemingly mostly English, is not fully known.} language pairs.
  
Overall, all these studies contribute to a better understanding of the abilities of instruction-based MT, and provide complementary angles, with variation across tasks, domains, language pairs, settings (e.g.\ context-aware MT or translation-memory-based MT), as well as evaluation metrics (BLEU, BLEURT, COMET) and protocols. In comparison, ours brings some additional observations related to MT performance across model sizes and for a large number of language pairs, as well as a new task (multilingual conversations).
}

Multilingual MT is also the subject of dedicated (monotask) architectures and training regimes. Originally introduced in \citep{dong-etal-2015-multi,firat-etal-2016-multi,luong-etal-2016-multitask} with limited language coverage, the latest versions of these approaches are able to handle hundreds of languages, including very low-resource language pairs \citep{fan-etal-2021-beyond,bapna-etal-2022-building,costa-etal-2022-nllb}. Although we found that \bloom is able to match this performance, given sufficient training data, we also see that it still lags behind for many languages pairs that are under-represented in its training data. 

\begin{table*}[!ht]
    \centering\small
    \resizebox{\textwidth}{!}{
    \begin{tabular}{llll}
    \toprule
    & Prompt name & Prompt & Target \\
      \midrule
       1--2 & \texttt{a\_good\_translation} & Given the following source text \textcolor{red}{(in L1)}: \textcolor{blue}{[source sentence]}, a good L2 translation is: & \textcolor{OliveGreen}{[target sentence]}\\
      3 & \texttt{version} & If the original version says \textcolor{blue}{[source sentence]} then the L2 version should say: & \textcolor{OliveGreen}{[target sentence]} \\
      4 & \texttt{gpt3} & What is the L2 translation of the sentence: \textcolor{blue}{[source sentence]}? & \textcolor{OliveGreen}{[target sentence]} \\ 
      5--6 & \texttt{xglm} &  \textcolor{red}{(L1:)} \textcolor{blue}{[source sentence]} = L2: & \textcolor{OliveGreen}{[target sentence]} \\
      7 & \texttt{translate\_as} & \textcolor{blue}{[source sentence]} translates into L2 as: & \textcolor{OliveGreen}{[target sentence]} \\
      \bottomrule
    \end{tabular}}
    \caption{\label{tab:prompt-examples}Seven MT prompts for the WMT'14 dataset \protect\cite{bojar-etal-2014-findings}. All prompts specify the target language (L2). Each prompt exists in a `target-only' version (\texttt{-target}), where only the target language is specified, and two prompts also exist in a second \texttt{-source+target} version, where the source language (in red and in brackets) is explicit in the instruction.}
\end{table*}
\section{\bloom Language Model}

\bloom is a large open-access multilingual model trained on 46 natural languages developed within the BigScience project \citep{lescao-etal-2022-bloom}. It is an auto-regressive language model designed to generate text to complete a user-entered text prefix, known as a prompt. It can be used for multiple tasks, including MT, question answering, etc. \bloom was trained on 1.6TB of text (of which 30\% English), from various sources, although 38\% of the data, known as the ROOTS corpus \citep{roots-corpus},\footnote{The ROOTS corpus can now be queried using the dedicated search tool \url{https://hf.co/spaces/bigscience-data/roots-search}.} is from Oscar web data \citep{OrtizSuarezSagotRomary2019}. The model is openly released on HuggingFace in multiple sizes, ranging from 560M to 176B parameters.\footnote{\url{https://hf.co/bigscience/bloom}}

\section{Evaluating \bloom on the MT task}\label{sec:mt}

\subsection{MT Datasets Used}
We experiment with three datasets, chosen to test different aspects of \bloom for MT: WMT \citep{bojar-etal-2014-findings}, Flores-101 \citep{goyal-etal-2022-flores} and DiaBLa \citep{bawden-etal-2020-diabla}.
We use the WMT 2014 news test sets %
for English$\leftrightarrow$French and English$\leftrightarrow$Hindi, which we take as representative high- and lower-resource language pairs with respect to \bloom's training data.\footnote{English, French and Hindi make up 30\%, 12.9\% and 0.7\% of the training data respectively \citep{roots-corpus}.} These test sets are somewhat outdated \citep{garcia-etal-2023-unreasonable}, but have been used repeatedly in past LLM evaluations and are included as standard benchmarks for comparison.
Flores-101 %
is a multi-parallel dataset in 101 languages, translated from original English sentences.\ifthenelse{\boolean{long}}{%
In fact, evaluations into English are bound to yield overly good results (e.g.\ \citep{toral-etal-2018-attaining}) and between other languages may mostly reflect their similarity with the original English.}{} We use it to test and compare \bloom's multilinguality, including for low-resource languages.%
\ifthenelse{\boolean{long}}{\footnote{An extended version, Flores-200, has been recently released \citep{costa-etal-2022-nllb}, which is larger and covers approximately twice as many languages. As this new version was released late in our evaluation process and had only been used in one paper, we decided to stick to Flores-101.}}{}
DiaBLa is a bilingual test set of spontaneous written dialogues between English and French speakers, mediated by MT. We use this as a test of MT in an informal domain and the impact of (cross-lingual) linguistic context in MT.

\subsection{Experimental setup \label{ssec:experiments}}

We evaluate and compare \bloom (and its variants) using the Language Model Evaluation Harness \citep{eval-harness} %
in 0-shot and few-shot settings. For few-shot, $k$ examples are prefixed to the prompt and separated with \#\#\# as shown in Example~\ref{ex:few-shot-prompt} (1-shot example is underlined).

\enumsentence{\label{ex:few-shot-prompt}\small \textbf{Input}: \underline{French: je m'ennuie = English: I'm bored.} \#\#\# English: Is that your dog that's just wandered in over there? = French:\\
\textbf{Reference}: Est-ce que c'est votre chien qui vient de rentrer par là~? }

Results are reported on the datasets' test splits. Few-shot examples are randomly taken from the data splits according to availability (train for WMT, dev for Flores-101 and test for DiaBLa).
We evaluate using BLEU \citep{papineni-etal-2002-bleu} as implemented in SacreBLEU \citep{post-2018-call}, using as tokenisation \texttt{13a} for WMT and DiaBLa and \texttt{spm} for Flores-101 as recommended \citep{costa-etal-2022-nllb}.\footnote{BLEU+case:mixed+smooth.exp+\{13a,spm\}+version.2.2.1} BLEU has many shortcomings but is good enough to provide quantitative comparisons for most systems used in this study.
We additionally use COMET \citep{rei-etal-2020-comet} for finer grained comparisons when the scores are closer.

\subsubsection{Comparative models}
In our cross-dataset comparison (Section~\ref{sec:dataset-comparison}), we compare \bloom to other LLMs: (i)~two task-fine-tuned models: \tzero\footnote{\url{https://hf.co/bigscience/T0}} \citep{sanh-etal-2022-multitask}, trained on English texts, and \mtzero{}\footnote{\url{https://hf.co/bigscience/mt0-xxl}} \citep{muenninghoff-etal-2022-bloomz}, the multilingual version, and (ii)~\opt{}\footnote{\url{https://hf.co/facebook/opt-66b}} \citep{opt}, an English generative LM. We evaluate all models on the same prompt \texttt{xglm-source+target}. To evaluate multiple language pairs with Flores-101, we compare (as a topline) to the supervised 615M-parameter MT model M2M-100 \citep{fan-etal-2021-beyond}, using the scores computed by \citet{goyal-etal-2022-flores}.

\subsubsection{Prompts}\label{sec:prompts}
We use several prompts, designed to illustrate different sources of variation: (i)~the inclusion (or not) of the source language name, %
(ii)~the relative order of source and target language names, (iii)~the position of the source sentence (beginning or end of the prompt) and (iv)~the prompt's verbosity. These prompts, available in PromptSource \citep{bach-etal-2022-promptsource}, are shown in Table~\ref{tab:prompt-examples}.
The first three are inspired by previous work:\footnote{This was not always straightforward due to incomplete documentation concerning (a)~prompts tested, and (b)~those actually used in each experiment (e.g.\ different ones for 0-shot and few-shot runs \citep{chowdhery-etal-2022-palm}).} \citep{gpt3} for \texttt{gpt3},\ifthenelse{\boolean{long}}{\footnote{Used only it seems, for zero-shot learning  in the form ``Q: what is the L2 translation of sentence [source sentence]. A:'', where special tokens Q and A are the query and the answer texts (cf.\ Figure G.36, pp~59).}}{} \citep{lin2021xglm} for \texttt{xglm} and \citep{wei-etal-2022-flan} for \texttt{translate\_as}, which also resembles \citet{raffel-etal-2020-exploring}'s prompt (\textsl{Translate English to German: ``[source text]'': [target sentence]})\ifthenelse{\boolean{long}}{, also used in \citep{wei-etal-2022-flan,garcia-firat-2022-using}.}{.}

\ifthenelse{\boolean{long}}{
  Considering the entries in Table~\ref{tab:prompt-examples}, we can see that ``prompting'' in fact refers to two distinct aspects of the input: (i)~the formulation of the task in natural language and (ii)~the presentation of related examples (for few-shot setups) interleaved with language tags (perhaps more clearly referred to as \textit{priming} by \citet{pham-etal-2020-priming}). As illustrated by the \texttt{xglm} prompt for example, the instruction part can reduced to one single word.
  As our results below suggest, the instruction mostly matters in 0-shot setups, but can almost be dispensed with in few-shot scenarios. The authors of \citep{gpt3} and \citep{hendy-etal-2023-howgood} also use a verbose, instruction-like prompt in their zero-shot setup, and a much more compact one for few shots experiments.
Also note that InstructGPT's prompt combines both an instruction and language tags \cite[p.~49]{long-etal-2022-training}.}{}

\section{Evaluation results \label{sec:results}}

Our evaluation of \bloom starts with a comparison across the three datasets and detection of major MT errors with a focus on WMT (Section~\ref{sec:dataset-comparison}) and then we present more in-depth analyses of particular aspects: (i)~using WMT, a comparative study of \bloom model sizes (Section~\ref{sec:model-size}) and prompts (Section~\ref{sec:per-prompt}), (ii)~using Flores-101 an evaluation of more language pairs and cross-lingual few-shot transfer (Section~\ref{sec:flores}), and (ii)~using DiaBLa, a study of the use of linguistic context (Section~\ref{sec:context}). 
 
\subsection{Comparison across datasets}\label{sec:dataset-comparison}

\begin{table}[!ht]
\begin{subtable}[h]{0.48\textwidth}
            \centering\small
            \resizebox{\linewidth}{!}{
            \begin{tabular}{lrrrrrrrr}
            \toprule
            & \multicolumn{4}{c}{0-shot} & \multicolumn{4}{c}{1-shot} \\
            & \bloom & T0 & mT0 & OPT & \bloom & T0 & mT0 & OPT \\
            \midrule
\multicolumn{5}{l}{WMT 2014} \\
\midrule
en$\rightarrow$fr & 14.9 & 1.2 & 29.3 & 12.9 & 27.8 & 1.4 & 25.2 & 21.9 \\
fr$\rightarrow$en & 15.5 & 25.8 & 32.9 & 15.5 & 34.6 & 21.0 & 30.0 & 24.6 \\
en$\rightarrow$hi & 6.8 & 0.2 & 11.2 & 0.1 & 13.6 & 0.1 & 9.5 & 0.1 \\
hi$\rightarrow$en & 12.1 & 0.0 & 26.1 & 0.4 & 25.0 & 0.0 & 20.1 & 0.6 \\
\midrule
\multicolumn{5}{l}{DiaBLa} \\
\midrule
en$\rightarrow$fr & 0.9 & 0.5 & 28.4 & 0.5 & 5.7 & 0.6 & 21.0 & 15.5 \\
fr$\rightarrow$en & 0.8 & 25.5 & 35.0 & 0.8 & 12.1 & 20.6 & 26.9 & 12.1 \\
\midrule
\multicolumn{5}{l}{Flores-101} \\
\midrule
en$\rightarrow$fr & 2.8 & 1.9 & 55.5 & 2.8 & 45.0 & 2.1 & 53.5 & 24.4 \\
fr$\rightarrow$en & 2.7 & 31.9 & 60.1 & 2.6 & 45.6 & 24.9 & 58.2 & 16.7 \\
en$\rightarrow$hi & 1.3 & 0.1 & 67.7 & 0.1 & 27.2 & 0.1 & 54.7 & 0.1 \\
hi$\rightarrow$en & 3.4 & 0.0 & 59.5 & 0.1 & 35.1 & 0.2 & 57.3 & 0.5 \\
\bottomrule
\end{tabular}}
\caption{\label{tab:xglm-main-orig}
Original predictions}
\end{subtable}
\begin{subtable}[h]{0.48\textwidth}
            \centering\small
            \resizebox{\linewidth}{!}{
            \begin{tabular}{lrrrrrrrrr}
            \toprule
            & \multicolumn{4}{c}{0-shot} & \multicolumn{4}{c}{1-shot} \\
            & \bloom & T0 & mT0 & OPT & \bloom & T0 & mT0 & OPT \\
            \midrule
\multicolumn{5}{l}{WMT 2014} \\
\midrule
en$\rightarrow$fr & 32.2 & 1.2 & 29.2 & 18.9 & 36.3 & 1.4 & 25.2 & 22.3 \\
fr$\rightarrow$en & 37.2 & 25.8 & 32.9 & 33.2 & 38.2 & 21.1 & 29.9 & 33.2 \\
en$\rightarrow$hi & 12.1 & 0.2 & 11.2 & 0.1 & 15.7 & 0.1 & 9.5 & 0.1 \\
hi$\rightarrow$en & 24.3 & 0.0 & 26.1 & 0.5 & 25.0 & 0.0 & 20.1 & 0.6 \\
\midrule
\multicolumn{5}{l}{DiaBLa} \\
\midrule
en$\rightarrow$fr & 24.2 & 0.5 & 28.4 & 17.4 & 37.6 & 0.6 & 21.9 & 20.7 \\
fr$\rightarrow$en & 22.9 & 25.5 & 34.9 & 36.8 & 41.4 & 21.1 & 27.2 & 37.6 \\
\midrule
\multicolumn{5}{l}{Flores-101} \\
\midrule
en$\rightarrow$fr & 26.9 & 1.9 & 55.3 & 21.4 & 49.3 & 2.1 & 53.4 & 28.4 \\
fr$\rightarrow$en & 40.3 & 31.9 & 60.0 & 39.4 & 47.2 & 25.2 & 58.2 & 39.8 \\
en$\rightarrow$hi & 7.7 & 0.1 & 67.7 & 0.1 & 29.5 & 0.1 & 54.7 & 0.1 \\
hi$\rightarrow$en & 30.2 & 0.0 & 59.5 & 0.2 & 35.1 & 0.2 & 57.3 & 0.5 \\
\bottomrule
\end{tabular}}
\caption{\label{tab:xglm-main-truncated}
Truncated predictions}
\end{subtable}
\caption{\label{tab:xglm-main-comparison}Cross-dataset comparison of BLEU scores (spBLEU for Flores-101) using the \texttt{xglm-source+target} prompt.}
\end{table}

We first prompt \bloom and the comparative models using the same prompt across datasets, %
restricting the directions tested to en$\leftrightarrow$fr and to en$\leftrightarrow$hi. We choose to systematically use the \texttt{xglm-source+target} prompt (Table~\ref{tab:prompt-examples}), %
which corresponds to the following template: %

\enumsentence{\label{ex:xglm-s+t}\small\texttt{L1: [source sentence] = L2:}} 
where \texttt{L1} and \texttt{L2} refer to the source and target languages respectively (e.g.~English and French for en$\rightarrow$fr) and \texttt{[source sentence]} is replaced by a given source sentence. 

BLEU scores are in Table~\ref{tab:xglm-main-orig} for both 0-shot and 1-shot (results with COMET are given in Appendix~\ref{app:comet-results}).
There are issues for 0-shot MT for all directions, particularly when translating into non-English languages, (BLEU scores are systematically lower than into English). Even into English, the scores remain low with respect to state of the art (e.g.~2.7 BLEU for Flores-101 fr$\rightarrow$en \bloom vs.\,60.1 for \mtzero).%
\footnote{
For comparison, \citep{palm} reports state-of-the art BLEU scores for supervised MT as 45.6 and 45.4 for WMT14 en$\rightarrow$fr and fr$\rightarrow$en respectively.
} 
\mtzero's good results illustrate the benefits of prompt fine-tuning. Results are considerably improved in the 1-shot setting (except for \tzero{} and \mtzero, both fine-tuned) and are comparable to other LLM MT results reported in the literature; they are markedly better for \bloom than for \opt, which was mostly trained on English data. These general observations apply for WMT and Flores-101, while issues remain for DiaBLa. 

\paragraph{What are the main issues?}
We identify two main issues (particularly for 0-shot): (i)~over-generation and (ii)~generating in the wrong language (an obvious problem for MT). Examples~\ref{ex:flores-0-shot} and~\ref{ex:flores-0-shot-chatty} illustrate some instances of these problems. The first example shows a very common problem, with repetition of the prompt in multiple languages until the maximum generation length is reached.

\enumsentence{\label{ex:flores-0-shot}\small \textbf{Input:} French: Il dit avoir conçu une sonnette de porte Wi-Fi. = English: \\
\textbf{\bloom:} He says he conceived a Wi-Fi doorbell.\,\textcolor{red}{=\,Spanish: Dice que concibió un timbre de puerta Wi-Fi. = German: Er sagt, er hat einen WLAN-Türglocken gegründet. = Italian: Dice di aver concepito un chiuso Wi-Fi [\ldots]}}

\enumsentence{\label{ex:flores-0-shot-chatty}\small
\textbf{Input:} French: Le tigre fait partie de la même famille (genre Panthera) que les lions, les léopards et les jaguars. Ces quatre félins sont les seuls capables de rugir. = English:	\\
\textbf{\bloom:} The tiger is a member of the same family (genus Panthera) as lions, leopards and jaguars. These four cats are the only ones capable of roaring. \textcolor{red}{Le tigre est un grand félin qui vit en Asie. Il est le plus grand des félins. Il est aussi le plus gros carnivore terrestre. Il est très rapide et peut courir à plus de 60 km/h. [\dots]}} %

\paragraph{Separating MT quality from overgeneration}

Overgeneration as seen in Example~\ref{ex:flores-0-shot} is a separate issue from \bloom's capacity to translate into another language. %
We therefore devise a custom truncating method for this type of overgeneration such that only the first translation in a prediction is kept, i.e.~anything after a newline or the regular expression pattern \texttt{= .+?:} is discarded. 

Results after truncation (Table~\ref{tab:xglm-main-truncated}) show that for all three datasets, 0-shot and 1-shot scores are significantly improved %
(e.g.~1-shot DiaBLa fr$\rightarrow$en increases from 12.05 to 41.36 and 0-shot Flores-101 hi$\rightarrow$en increases from 3.40 to 30.19). \bloom is capable of performing good MT but has a problem knowing when to stop generating. We use the same truncation elsewhere too and indicate when we show results for original or truncated outputs.

\begin{table}[!ht]
  \centering\small
  \resizebox{\linewidth}{!}{
  \begin{tabular}{l*{8}{r}}
  \toprule
    & \multicolumn{2}{c}{en$\rightarrow$fr} & \multicolumn{2}{c}{fr$\rightarrow$en} & \multicolumn{2}{c}{en$\rightarrow$hi} & \multicolumn{2}{c}{hi$\rightarrow$en} \\
       & 0 & 1 & 0 & 1 & 0 & 1 & 0 & 1\\
    \midrule
    Target     & 2814 &2959&2954&2979    &1998&2431 & 2469 & 2499\\
    Source      &   181 &   32&   47 &    22 & 476 &  48  &     29 &      2\\
    Other        &      8  &   12&     2 &    2 &   33 &   28  &      9 &       6\\
    \midrule
    Total       & 3003 & 3003 & 3003 & 3003 & 2507 & 2507 & 2507 & 2507 \\ 
\bottomrule
  \end{tabular}}
  \caption{\label{tab:wrong_langid}%
  The number of outputs (after truncation) classified as being in the (correct) target language, the source language, or another language for 0-shot and 1-shot setups (for WMT).}
\end{table}

\paragraph{Detecting generation in the wrong language}

We automatically detect the language of predictions using fasttext langid\footnote{\url{https://fasttext.cc/docs/en/language-identification.html}, using the compressed version \texttt{lid.176.ftz}.} \citep{joulin-etal-2017-bag}. Table~\ref{tab:wrong_langid} shows the number of translations identified as being in the correct target language, or alternatively in the source or another language for 0-shot and 1-shot setups after truncation.\ifthenelse{\boolean{long}}{\footnote{Raw tables can be found in Tables~\ref{tab:langid_wmt14_enfr} and~\ref{tab:langid_wmt14_enhi} in Appendix~\ref{app:langid_overgen}.}}{\footnote{See the raw results in Tables~\ref{tab:langid_wmt14_enfr} and~\ref{tab:langid_wmt14_enhi} in Appendix~\ref{app:langid_overgen}.}}$^,$\footnote{These numbers are better than the initial ones reported in \citep{lescao-etal-2022-bloom}, as we use a different prompt and truncation. See below for a detailed analysis per prompt.} The number of sentences in the correct target language increases from 0- to 1-shot, particularly for the two non-English target languages. When translating into Hindi (0-shot), 1/5 (509) of predictions are not detected as Hindi; %
the 1-shot largely mitigates the issue (only 76 outputs are in the wrong language). 

\begin{figure}[!ht]

\begin{subfigure}[p]{0.245\textwidth}
\pgfplotsset{
  every axis plot/.append style={line width=0.9pt}
}
\begin{tikzpicture}
	\begin{axis}[
		height=3.8cm,  
		width=1.15\textwidth,
		grid=major,
		xmin=0,
		xmax=5,
        ymin=0,
		ymax=40.1,
		ylabel = {BLEU score},
		label style={font=\small},
		xlabel = {\#fewshot examples}, 
		xtick distance = 1,
		tick label style={font=\small},
        axis x line*=bottom,
        axis y line*=left,
        legend style={draw=none,
            at={(0.8,0.5), font=\small,
            /tikz/every even column/.append style={column sep=0.9cm}},
    },
    legend columns=4 %
	]
    \addlegendentry{en$\rightarrow$fr}
    \addplot coordinates {
        (0, 14.91) (1, 27.83) (2, 35.09) (5, 37.89)
	};
    \addlegendentry{fr$\rightarrow$en}
    \addplot coordinates {
        (0, 15.52) (1, 34.61) (2, 37.19) (5, 38.94)
	};
  \legend{};
\end{axis}
\end{tikzpicture}
\caption{\label{fig:increase-num-examples-orig}Original outputs.}
\end{subfigure}
\begin{subfigure}[p]{0.215\textwidth}
\pgfplotsset{
  every axis plot/.append style={line width=0.9pt}
}
\begin{tikzpicture}
	\begin{axis}[
		height=3.8cm,
		width=1.3\textwidth,
		grid=major,
		xmin=0,
		xmax=5,
        ymin=0,
		ymax=40.1,
        yticklabels={\phantom{0},\phantom{10},\phantom{20},\phantom{30}}\phantom{40}
		ylabel = {}, %
		label style={font=\small},
		xlabel = {\#fewshot examples}, 
		xtick distance = 1,
		tick label style={font=\small},
        axis x line*=bottom,
        axis y line*=left,
        legend style={draw=none,
            at={(1,0.5), font=\small,
            /tikz/every even column/.append style={column sep=0.9cm}},
    },
    legend columns=1 %
	]
    \addlegendentry{en$\rightarrow$fr}
    \addplot coordinates {
        (0, 32.25) (1, 36.29) (2, 37.62) (5, 37.88)
	};
    \addlegendentry{fr$\rightarrow$en}
    \addplot coordinates {
        (0, 37.16) (1, 38.18) (2, 38.31) (5, 38.98)
	};

\end{axis}
\end{tikzpicture}
\caption{\label{fig:increase-num-examples-truncated} Truncated outputs.}
\end{subfigure}
\caption{\label{fig:increase-num-examples-orig+truncated} BLEU scores for WMT 2014 en$\leftrightarrow$fr and the \texttt{xglm} prompt, with an increasing number of few-shot examples.}
\end{figure}
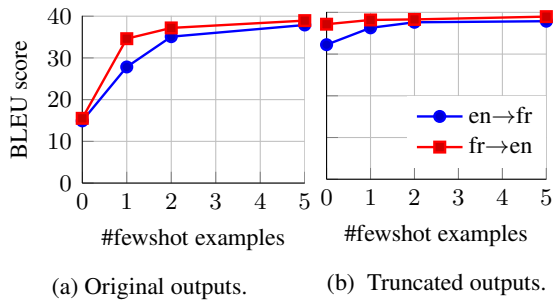

\paragraph{Increasing the number of few-shot examples}
Both problems improve significantly in the 1-shot setup, a trend that continues as the number of few-shot examples increases, resulting in higher BLEU scores, as can be seen in Figure~\ref{fig:increase-num-examples-orig+truncated} for WMT %
en$\leftrightarrow$fr. %
However, we see diminishing returns, particularly visible between 2 to 5 examples, suggesting that gains beyond 5-shot would be more marginal.

\subsection{\bloom model size}\label{sec:model-size}
Several versions of \bloom exist, with differing numbers of parameters. To test how size impacts performance, we report average scores and ranges for WMT across the seven prompts.
Table~\ref{tab:increase-bloom} shows that as the size decreases (from 176B to 560M parameters), the performance also decreases significantly. We see substantial gains for all models when moving from 0-shot to 1-shot, the smaller models (e.g. \bloom-7b1, \bloom-3b) slightly closing the gap with the largest one.
As the ranges in Table~\ref{tab:increase-bloom} are computed across prompts, we see that different prompts yield markedly different BLEU scores in the 0-shot setup; for 1-shot, we still see variations of 6-8 BLEU points between the best and the worst prompt. %
Similar analyses performed with post-processing and also for English$\leftrightarrow$Hindi \ifthenelse{\boolean{long}}{(Appendix~\ref{sec:permodel-appendix})}{(Appendix~\ref{sec:permodel-appendix})} confirm that (i)~truncation improves scores for all model sizes and prompts and (ii)~the choice of a bad prompt can result in catastrophic MT performance as compared to a good one.

\begin{table}[!ht]
\begin{subtable}[h]{0.48\textwidth}
\centering\small
\begin{tabular}{lrlrl}
\toprule
Model & \multicolumn{2}{c}{en$\rightarrow$fr} & \multicolumn{2}{c}{fr$\rightarrow$en} \\
\midrule
\bloom       & 11.2 & \scriptsize{3.0--22.0} & 15.4 & \scriptsize{10.3--26.8} \\
\bloom-7b1   & 6.5 & \scriptsize{1.5--12.1} & 12.8 & \scriptsize{4.8--25.1} \\
\bloom-3b    & 3.6 & \scriptsize{1.2--9.6} & 10.6 & \scriptsize{2.8--19.3} \\
\bloom-1b1   & 1.7 & \scriptsize{0.5--3.9} & 7.1 & \scriptsize{0.7--11.4}  \\
\bloom-560m  & 0.6 & \scriptsize{0.4--0.9} & 3.7 & \scriptsize{1.4--5.4} \\
\bottomrule
\end{tabular}
\caption{\label{tab:increase-bloom-0-shot}0-shot}
\end{subtable}
\hfill
\begin{subtable}[h]{0.48\textwidth}
\centering\small
\begin{tabular}{lrlrl}
\toprule
Model & \multicolumn{2}{c}{en$\rightarrow$fr} & \multicolumn{2}{c}{fr$\rightarrow$en} \\
\midrule
\bloom      &  32.6 & \scriptsize{27.8 --36.4} &  34.9 & \scriptsize{33.1--36.6} \\
\bloom-7b1  &  25.9 & \scriptsize{20.8--29.9}  &  29.1 & \scriptsize{25.4--32.5} \\
\bloom-3b   &  21.6 & \scriptsize{16.7--26.8}  &  25.7 & \scriptsize{18.6--29.6} \\
\bloom-1b1  &  10.1 & \scriptsize{6.3--13.2}   &  16.1 & \scriptsize{12.2--19.9} \\
\bloom-560m &   3.6 & \scriptsize{2.2--4.4}    &   8.6 &  \scriptsize{5.8--12.1} \\
\bottomrule
\end{tabular}
\caption{\label{tab:increase-bloom-1-shot}1-shot}
\end{subtable}
\caption{\label{tab:increase-bloom}Average BLEU scores and ranges across the seven prompts for decreasing sizes of \bloom (original outputs).}%
\end{table}

\subsection{Per-prompt analysis\label{sec:per-prompt}}

Looking at average WMT results computed with respect to prompt choice (using the prompts in Table~\ref{tab:prompt-examples}) allows us to further investigate cross-prompt variability. 

\paragraph{Which prompt works best?}

\begin{table*}[!ht]
\centering\small
\begin{tabular}{l*{4}{r@{ }c}}
\toprule
   &      \multicolumn{4}{c}{en$\rightarrow$fr} & 
               \multicolumn{4}{c}{fr$\rightarrow$en}  \\
  Prompt / Few-shot \# & \multicolumn{2}{c}{0} & \multicolumn{2}{c}{1} & \multicolumn{2}{c}{0} & \multicolumn{2}{c}{1} \\
\midrule 
  \texttt{a\_good\_translation-source+target}  &   6.7 & \scriptsize{0.6--15.4}  &  18.7  & \scriptsize{4.1--36.4}    &  11.0 & \scriptsize{5.4--14.2} &  25.8 & \scriptsize{11.6--36.6} \\
  \texttt{a\_good\_translation-target}   &   3.1 & \scriptsize{0.4--10.1}  &  20.3 & \scriptsize{3.2--35.5}    &  12.1 & \scriptsize{5.1--16.8} & \textbf{25.9} & \scriptsize{12.1--36.2} \\
  \texttt{gpt3-target}  &   2.5 & \scriptsize{0.5-- 7.9}   &  16.6 & \scriptsize{2.2--32.5}    &   4.5 & \scriptsize{0.7 --12.7}  &  19.3 &  \scriptsize{5.8--33.1} \\
  \texttt{translate\_as-target} &   3.3 & \scriptsize{0.4-- 5.0}   &  17.1 & \scriptsize{3.2--32.7}    &   6.9 & \scriptsize{2.1 --11.3}  &  21.6 &  \scriptsize{7.6--35.1} \\ 
\texttt{version-target} &   7.5 & \scriptsize{0.6--22.0}  &  \textbf{21.4} & \scriptsize{4.3--34.2}    &  \textbf{17.1} & \scriptsize{3.9--26.8}  &  24.9 &  \scriptsize{7.8--35.4} \\
\texttt{xglm-source+target} &   \textbf{8.3} & \scriptsize{0.9--14.9}  &  17.5 & \scriptsize{3.3-- 27.8}    &  11.8 & \scriptsize{5.0--15.5}  &  22.1 &  \scriptsize{7.8--34.6} \\
\texttt{xglm-target}   &   1.6 & \scriptsize{0.7-- 3.0}   &  16.7 & \scriptsize{4.4--29.0}    &   6.2 & \scriptsize{2.6--10.3}   &  20.7 &  \scriptsize{7.5--33.3} \\                                                      
\bottomrule
\end{tabular}
\caption{\label{tab:per-prompt-analysis-big-bloom-en-fr}Average, min and max BLEU scores by prompt for en$\leftrightarrow$fr (original outputs). Best average result per setting in bold.}
\end{table*}

 \begin{table*}[!ht]
 \centering\small
 \begin{tabular}{l*{4}{r@{ }c}}
 \toprule
    &   \multicolumn{4}{c}{en$\rightarrow$hi} & 
    \multicolumn{4}{c}{hi$\rightarrow$en}  \\
   Prompt / Few-shot \# & \multicolumn{2}{c}{0} & \multicolumn{2}{c}{1} & \multicolumn{2}{c}{0} & \multicolumn{2}{c}{1} \\
 \midrule
 \texttt{a\_good\_translation-source+target} & 0.7 & \scriptsize{0.1--1.9} &  5.8 & \scriptsize{0.3--14.5} & 4.8 & \scriptsize{0.9--10.2} &  13.1 & \scriptsize{2.8--24.6} \\
 \texttt{a\_good\_translation-target} & 0.2 & \scriptsize{0.1--0.8} & 5.5 & \scriptsize{0.3--14.1} &  6.3 & \scriptsize{1.1--13.0} &   13.2 & \scriptsize{ 2.8--24.8} \\
 \texttt{gpt3-target}    &  0.1 & \scriptsize{0.0--0.3}  &   1.4  & \scriptsize{ 0.0--6.5}  &  0.2 & \scriptsize{ 0.0--0.7} & 2.2 &   \scriptsize{ 0.0--10.0}  \\
 \texttt{version-target} &  0.7 & \scriptsize{0.1--2.0} &   5.6 & \scriptsize{ 0.2--14.0} &  \textbf{6.8} & \scriptsize{ 1.7--11.5} & \textbf{13.3} & \scriptsize{ 2.4--25.8} \\
 \texttt{xglm-source+target} &  \textbf{2.1} & \scriptsize{0.1--6.8} &  \textbf{6.9}  & \scriptsize{ 0.3--13.6} & 4.4 & \scriptsize{ 0.6--12.1} &  11.9 & \scriptsize{ 1.7--25.0} \\
 \texttt{xglm-target}    &  0.2 & \scriptsize{0.0--0.6} &  5.1   & \scriptsize{ 0.1--14.6} &  1.6 & \scriptsize{ 0.2--4.1}  &  6.6 & \scriptsize{ 0.5--13.2} \\
 \bottomrule
 \end{tabular}
  \caption{\label{tab:per-prompt-analysis-big-bloom-en-hi}Average, min and max BLEU scores per prompt for en$\leftrightarrow$hi (original outputs). Best average result per setting in bold.}
\end{table*}

This variability is illustrated in Tables~\ref{tab:per-prompt-analysis-big-bloom-en-fr} and \ref{tab:per-prompt-analysis-big-bloom-en-hi} report performance across prompts for en$\leftrightarrow$\{fr,hi\}, averaged over the five \bloom models from Section~\ref{sec:model-size}.\footnote{For a given prompt, the range mainly reflects the performance of the different sizes of \bloom model.} \ifthenelse{\boolean{long}}{The corresponding tables for truncated outputs are in Appendix~\ref{sec:perprompt-appendix}.}{The corresponding tables for truncated outputs are in Appendix~\ref{sec:perprompt-appendix}.}
\texttt{version} and \texttt{a\_good\_translation} (\texttt{source+target}) get the highest average (and maximum) scores. Both prompts are more verbose (instruction-like), but the performance gap in the 1-shot setting between these prompts and the simpler, `priming-style' prompts (e.g. \texttt{xglm}) narrows.
The worst results are seen for \texttt{gpt3}. With this prompt, 
translating into French after a text that only contains English seems particularly difficult: half of the 0-shot translations for \texttt{gpt3} are classified as non-French by langid (most of them are English). When translating into Hindi, only 10~outputs are detected as being in Hindi.

\paragraph{Does it help to specify the source language in the prompt?}
We compare the two versions (\texttt{-target} and \texttt{-source+target}) of \texttt{a\_good\_translation} and \texttt{xglm}. %
Results in Tables~\ref{tab:per-prompt-analysis-big-bloom-en-fr} and \ref{tab:per-prompt-analysis-big-bloom-en-hi} are inconclusive. For these language directions and prompts, we see small differences for 1-shot, which may be due to variance between runs. For 0-shot, it clearly helps  \texttt{xglm} to indicate the source language, but for the more verbose  \texttt{a\_good\_translation}, it helps one direction and hurts the other. This question would need to be further explored to draw more solid conclusions, including with non-English prompts.

\subsection{Evaluating more language directions \label{sec:flores}}

We further explore more language directions in the 1-shot setting using Flores-101. As in Section~\ref{sec:dataset-comparison}, we use the \texttt{xglm-source+target} prompt.\ifthenelse{\boolean{long}}{\footnote{It behaved well on average in the previous experiments and is one of the least verbose, making it more suitable in a multilingual setting.}}{}

\subsubsection{Per-language results}\label{sec:per-lang-results}

To optimise computational resources, instead of running all language combinations, we concentrate on: %
(i)~high-resource language pairs, (ii)~high$\rightarrow$mid-resource language pairs, (iii)~low-resource language pairs and (iv)~related languages (specifically Romance languages). Results are shown in Tables~\ref{tab:flores_summary-high-midhigh-results} and~\ref{tab:flores-results-low-romance} for original outputs, given that overgeneration is less problematic for 1-shot.

\paragraph{High-resource and high$\rightarrow$mid-resource}
The results for high-resource and high$\rightarrow$mid-resource language directions are generally good, surpassing M2M scores for high-resource, except for es$\rightarrow$fr.\footnote{%
French and Spanish, although related and comparably represented in ROOTS, have very different scores. Our preliminary analysis suggests that this is due to %
the Spanish references being less literal than the French\ifthenelse{\boolean{long}}{ and structurally more different from the original English}{}. \ifthenelse{\boolean{long}}{See Appendix~ \ref{sec:spanish-in-flores} for some examples.}{See Appendix~ \ref{sec:spanish-in-flores} for some examples.}}
This suggests that \bloom a has good multilingual capacity, even across scripts (between (extended) Latin, Chinese, Arabic and Devanagari scripts).

\paragraph{Low-resource}
For low-resource languages, the results are more variable; some language directions see better results than M2M, notably most into-English directions, but others are less good (e.g.~into Hindi and Swahili). Results for the lowest-resourced languages tested (sw$\leftrightarrow$yo and en$\leftrightarrow$yo) are particularly disappointing because the scores indicate that the resulting translations are meaningless, even though Yoruba and Swahili are present (although under-represented) in BLOOM's training data ($<$50k tokens each). 

\paragraph{Romance languages}
This contrasts with the results %
between Romance languages, where results are good across-the-board, including from and into Italian (it) and Galician (gl), which are not officially in the training data. Note that Galician shares many similarities with the other Romance languages, in particular with Portuguese (pt). These contrasted results show the performance of an LLM not only depends on the amount of training data, but also largely on the similarity with seen languages. To be complete, these analyses should also take into account the possibility of mislabellings in the training data,\footnote{In a personal communication, N.~Muennighoff estimates that Italian accounts for $\sim$0.33\% of the ROOTS corpus, slightly below the proportion of Hindi texts (0.47\%).} %
which have been found to explain a great deal of cross-lingual abilities of LLMs \citep{blevins-zettlemoyer-2022-language}.

\begin{table}[htbp]
\begin{subtable}[b]{0.48\textwidth}
\centering\small
\scalebox{0.86}{
\begin{tabular}{llrrrrr}
\toprule
Src $\downarrow$ & Trg $\rightarrow$ & ar & en & es & fr & zh \\
\midrule
\multirow{2}{*}{ar} & \bloom & -- & 40.3 & 23.3 & 33.1 & 17.7 \\
 & M2M & -- & 25.5 & 16.7 & 25.7 & 13.1 \\
\midrule
\multirow{2}{*}{en} & \bloom & 28.2 & -- & 29.4 & 45.0 & 26.7 \\
 & M2M & 17.9 & -- & 25.6 & 42.0 & 19.3 \\
\midrule
\multirow{2}{*}{es} & \bloom & 18.8 & 32.7 & -- & 24.8 & 20.9 \\
 & M2M & 12.1 & 25.1 & -- & 29.3 & 14.9 \\
\midrule
\multirow{2}{*}{fr} & \bloom & 23.4 & 45.6 & 27.5 & -- & 23.2 \\
 & M2M & 15.4 & 37.2 & 25.6 & -- & 17.6 \\
\midrule
\multirow{2}{*}{zh} & \bloom & 15.0 & 30.5 & 20.5 & 26.0 & -- \\
 & M2M & 11.6 & 20.9 & 16.9 & 24.3 & -- \\
\bottomrule
\end{tabular}}
\caption{High-resource language pairs.}
\label{tab:flores101_summary:high-high}

\scalebox{0.86}{
\begin{tabular}{llrrrrr}
\toprule
Src $\downarrow$ & Trg $\rightarrow$ & en & fr & hi & id & vi \\
\midrule
\multirow{2}{*}{en} & \bloom & -- & 45.0 & 27.2 & 39.0 & 28.5 \\
 & M2M & -- & 42.0 & 28.1 & 37.3 & 35.1 \\
\midrule
\multirow{2}{*}{fr} & \bloom & 45.6 & -- & 18.5 & 31.4 & 32.8 \\
 & M2M & 37.2 & -- & 22.9 & 29.1 & 30.3 \\
\midrule
\multirow{2}{*}{hi} & \bloom & 35.1 & 27.6 & -- & -- & -- \\
 & M2M & 27.9 & 25.9 & -- & -- & -- \\
\midrule
\multirow{2}{*}{id} & \bloom & 43.2 & 30.4 & -- & -- & -- \\
 & M2M & 33.7 & 30.8 & -- & -- & -- \\
\midrule
\multirow{2}{*}{vi} & \bloom & 38.7 & 26.8 & -- & -- & -- \\
 & M2M & 29.5 & 25.8 & -- & -- & -- \\
\bottomrule
\end{tabular}}
\caption{High$\rightarrow$mid-resource language pairs.}
\label{tab:flores101:high-mid}
\end{subtable}
\caption{\label{tab:flores_summary-high-midhigh-results}1-shot MT results (spBLEU) on the FLORES-101 devtest set (original outputs).}
\end{table}

\begin{table}[!ht]
 \begin{subtable}[t]{0.48\textwidth}
 \centering\small
\scalebox{0.86}{
\begin{tabular}{llrrrrr}
\toprule
Src$\downarrow$ & Trg$\rightarrow$ & en & bn & hi & sw & yo \\
\midrule
\multirow{2}{*}{en} & \bloom & -- & 24.6 & 27.2 & 20.5 & 2.6 \\
 & M2M & -- & 23.0 & 28.1 & 26.9 & 2.2 \\
\midrule
\multirow{2}{*}{bn} & \bloom & 29.9 & -- & 16.3 & -- & -- \\
 & M2M & 22.9 & -- & 21.8 & -- & -- \\
\midrule
\multirow{2}{*}{hi} & \bloom & 35.1 & 23.8 & -- & -- & -- \\
 & M2M & 27.9 & 21.8 & -- & -- & -- \\
\midrule
\multirow{2}{*}{sw} & \bloom & 37.4 & -- & -- & -- & 1.3 \\
 & M2M & 30.4 & -- & -- & -- & 1.3 \\
\midrule
\multirow{2}{*}{yo} & \bloom & 4.1 & -- & -- & 0.9 & -- \\
 & M2M & 4.2 & -- & -- & 1.9 & -- \\
\bottomrule
\end{tabular}}
\caption{Low-resource languages}
\label{tab:flores101_summary:high-low}

\resizebox{\linewidth}{!}{
\begin{tabular}{llrrrrrr}
\toprule
Src$\downarrow$ & Trg$\rightarrow$ & ca & es & fr & gl & it & pt \\
\midrule
\multirow{2}{*}{ca} & \bloom & -- & 28.9 & 33.8 & 19.2 & 19.8 & 33.0 \\
 & M2M & -- & 25.2 & 35.1 & 33.4 & 25.5 & 35.2 \\
\midrule
\multirow{2}{*}{es} & \bloom & 31.2 & -- & 24.8 & 23.3 & 16.5 & 29.1 \\
 & M2M & 23.1 & -- & 29.3 & 27.5 & 23.9 & 28.1 \\
\midrule
\multirow{2}{*}{fr} & \bloom & 37.2 & 27.5 & -- & 24.9 & 24.0 & 38.9 \\
 & M2M & 28.7 & 25.6 & -- & 32.8 & 28.6 & 37.8 \\
\midrule
\multirow{2}{*}{gl} & \bloom & 37.5 & 27.1 & 33.8 & -- & 18.3 & 32.2 \\
 & M2M & 30.1 & 27.6 & 37.1 & -- & 26.9 & 34.8 \\
\midrule
\multirow{2}{*}{it} & \bloom & 31.0 & 25.4 & 31.4 & 20.2 & -- & 29.2 \\
 & M2M & 25.2 & 29.2 & 34.4 & 29.2 & -- & 31.5 \\
\midrule
\multirow{2}{*}{pt} & \bloom & 39.6 & 28.1 & 40.3 & 27.1 & 20.1 & -- \\
 & M2M & 30.7 & 26.9 & 40.2 & 33.8 & 28.1 & -- \\
\bottomrule
\end{tabular}}
\caption{Romance languages}
\label{tab:flores101_summary:same-family}
\end{subtable}
\caption{\label{tab:flores-results-low-romance}1-shot MT results (spBLEU) on the Flores-101 devtest set (original outputs). }
\end{table}

\begin{table}[!ht]
\centering\small
\resizebox{\linewidth}{!}{
\begin{tabular}{llrrrr}
\toprule
                &  & \multicolumn{2}{c}{Original} & \multicolumn{2}{c}{Truncated} \\
\multicolumn{2}{l}{1-shot example direction type} & spBLEU & COMET & spBLEU & COMET \\
\midrule
Same & bn$\rightarrow$en & 29.9 & 0.444 & 29.9 & 0.444 \\
Opposite & en$\rightarrow$bn & 21.8 & 0.313 & 29.4 & 0.414 \\
\midrule
Related src & hi$\rightarrow$en & 30.1 & 0.449 & 30.5 & 0.460 \\
Related src (WMT) & hi$\rightarrow$en & 29.1 & 0.422 & 29.1 & 0.427 \\
HR unrelated src & fr$\rightarrow$en & 17.2 & 0.315 & 29.7 & 0.396 \\
HR unrelated src & fr$\rightarrow$ar & 8.4 & -0.102 & 28.0 & 0.322 \\
\bottomrule
\end{tabular}}
\caption{\label{tab:bn-en-fewshot-variation}1-shot results for Flores bn$\rightarrow$en when varying the language direction of 1-shot examples. HR=high-resource.}
\end{table}

\subsubsection{Cross-lingual transfer}\label{sec:transfer}
1-shot results are positive for many of the language directions tested (including low-resource), provided they are sufficiently represented in the ROOTS corpus. To better understand how cross-lingual \bloom is and how the 1-shot mechanism functions, we vary the language direction of the few-shot examples, taking Bengali$\rightarrow$English (bn$\rightarrow$en) translation as our case study. Taking random 1-shot dev set examples,\footnote{The random seed is kept the same for all runs.} we compare the use of 1-shot examples from (i)~the same direction (bn$\rightarrow$en), (ii)~the opposite direction (en$\rightarrow$bn), (iii)~a language direction whereby the source languages are related (hi$\rightarrow$en), (iv)~the same related direction but from a different dataset (the WMT dev set) (v)~a high-resource direction into the same target language (fr$\rightarrow$en) and (vi)~a high-resource unrelated language direction (fr$\rightarrow$ar).

The results (Table~\ref{tab:bn-en-fewshot-variation}) %
show that cross-lingual transfer is possible, but using a different language direction can impact  overgeneration and translation quality. %
The unrelated direction fr$\rightarrow$ar gives the worst results, with most overgeneration (see the score difference between original and truncated), but also the worst quality after truncation, suggesting that language relatedness does play a role.
Overgeneration is still a problem (although less so) when using the opposite direction (en$\rightarrow$bn) or the same target language (fr$\rightarrow$en). Using a related (higher-resource) source language (hi$\rightarrow$en) reduces overgeneration and also gives the best MT results. %
However, better results are seen when using Flores-101 rather than WMT examples, suggesting that in-domain examples are best.

\subsection{Use of Linguistic Context}\label{sec:context}

\begin{table}[!ht]
        \centering\small
        \resizebox{\linewidth}{!}{
        \begin{tabular}{llcrrrr}
        \toprule
        \multicolumn{2}{c}{1-shot example} & & \multicolumn{2}{c}{en$\rightarrow$fr} & \multicolumn{2}{c}{fr$\rightarrow$en} \\
        Origin & Dir. & Trunc. & BLEU & COMET & BLEU & COMET \\
        \toprule
\multirow{2}{*}{Rand.} & \multirow{2}{*}{rand.} & $\times$ & 5.7 & 0.342 & 12.1 & 0.614 \\
&  & $\checkmark$ & 37.6 & 0.634 & 41.4 & 0.758 \\
\midrule
\multirow{2}{*}{Prev.} & \multirow{2}{*}{rand.} &  $\times$ & 6.1 & 0.328 & 12.3 & 0.617 \\
 & & $\checkmark$ & 38.5 & 0.614 & 41.6 & 0.751 \\
\midrule
\multirow{2}{*}{Prev.} & \multirow{2}{*}{same} & $\times$ & 19.3 & 0.597 & 20.7 & 0.719 \\
&  & $\checkmark$ & \textbf{39.0} & \textbf{0.632} & \textbf{42.1} & \textbf{0.761} \\
\midrule
\multirow{2}{*}{Prev.} & \multirow{2}{*}{opp.} & $\times$ & 3.6 & 0.064 & 8.6 & 0.518 \\
& & $\checkmark$ & 37.8 & 0.590 & 41.2 & 0.742 \\
\bottomrule
        \end{tabular}}
        \caption{Comparison of 1-shot results (BLEU) for DiaBLa when using the previous/random sentence for the 1-shot example (using the \texttt{xglm-source+target} prompt). In bold are the best results for each language direction.}
        \label{tab:diabla-context-results}
        \end{table}

There has been a considerable amount of research on linguistic context in MT, e.g.~to  disambiguate lexically ambiguous texts or when additional information is necessary for the output to be well-formed (e.g.~translating anaphoric pronouns into a language that requires agreement with a coreferent) \citep{Hardmeier2012-discourse,libovicky-helcl-2017-attention,bawden-etal-2018-evaluating,voita-etal-2018-context,lopes-etal-2020-document,Nayak2022-investigating}.

We test the usefulness of linguistic context in DiaBLa in the 1-shot setting (again using \texttt{xglm-source+target}) by changing the origin of 1-shot examples: (i)~a random example vs. (ii)~the previous dialogue utterance. If linguistic context is useful, we would expect there to be an improvement for (ii). 
We also vary the language direction of the 1-shot example. By default, given that the dataset is bilingual, the direction of 1-shot examples is en$\rightarrow$fr or fr$\rightarrow$en, independent of the current example's direction. Given the results in Section~\ref{sec:transfer} and the poor 0-shot results in Table~\ref{tab:xglm-main-orig}, it is important to account for this to provide a fair comparison. We therefore compare each type of context (random/previous) with (i)~the same random  directions, and (ii-iii)~the same (and opposite) language directions as the current example. %
We show results for original and truncated outputs.

Results are shown in Table~\ref{tab:diabla-context-results}. %
Truncation helps considerably;  even for 1-shot, \bloom struggles not to overgenerate and this is considerably reduced when the same rather than the opposite language direction is used for the 1-shot example.
It is unclear whether using previous rather than random context helps: BLEU is higher (38.5 vs.~37.6), whereas COMET is lower (0.328 vs.~0.342). %
These differences could be the result of randomness in 1-shot example selection, and different results could be obtained with a different random seed. %
Despite these inconclusive results, it is clear that using previous context influences the translation, for better or worse. For evidence of this, see Table~\ref{tab:diabla-examples} in Appendix~\ref{app:diabla-examples}, which provides three such examples: (i)~an unlucky negative influence on the translation of an ambiguous word \textit{glace} `ice cream or mirror' from the previous context, resulting in the wrong sense being chosen, (ii)~the use of a coreferent \textit{instrument} `instrument' from the previous sentence and (iii)~the correct gender agreement of the pronoun \textit{they} into French (\textit{elles} `they (fem.)' as opposed to \textit{ils} `they (masc.)') to correspond to the feminine coreferent \textit{filles} `girls'. %

\section{Conclusion}

We have evaluated \bloom's MT performance across three datasets and multiple language pairs. %
While there remain problems of overgeneration and generating in the wrong language (particularly for 0-shot MT), MT quality is significantly improved in few-shot settings, closer to state-of-the-art results. Low-resource MT remains challenging for some language pairs, despite the languages being in the training data, questioning what it means to be %
a \bloom language. However, we see evidence for cross-lingual transfer for non-\bloom languages and when using few-shot examples from other language pairs. Finally, although using linguistic context does not give improvements with automatic metrics, there is evidence that discursive phenomena are taken into account.

\section*{Acknowledgements}
This work was made possible with the collective efforts of the BigScience community, who designed, developed and prepared the tools and datasets used to train \bloom. Special mention to evaluation working group members %
and especially to Niklas Muenninghoff and Pawan Sasanka Ammanamanchi for producing some of our results.

This work was granted access to the HPC resources of Institut du développement et des ressources en informatique scientifique (IDRIS) du Centre national de la recherche scientifique (CNRS) under the allocations 2021-AD011011717R1, AD011012254R2, 2021-A0101012475 and 2022-AD010614012 made by Grand équipement national de calcul intensif (GENCI). 
R. Bawden's participation was partly funded by her chair position in the PRAIRIE institute, funded by the French national agency ANR as part of the ``Investissements d'avenir'' programme under the reference ANR-19-P3IA-0001, and by her Emergence project, DadaNMT, funded by
Sorbonne Université.

\ifthenelse{\boolean{long}}
{\bibliography{eamt23}}{\bibliography{eamt23-compressed}}

\ifthenelse{\boolean{long}}
{\clearpage
\appendix

\section{COMET Results for Main Comparison}\label{app:comet-results}

Table~\ref{tab:xglm-main-comparison-comet} shows the COMET scores for the cross-dataset and model comparison. The conclusions drawn for the Table~\ref{tab:xglm-main-comparison} with BLEU scores hold here.

\begin{table}[!ht]
\begin{subtable}[h]{0.48\textwidth}
            \centering\small
            \resizebox{\linewidth}{!}{
            \begin{tabular}{lrrrrrrrr}
            \toprule
            & \multicolumn{4}{c}{0-shot} & \multicolumn{4}{c}{1-shot} \\
            & \bloom & T0 & mT0 & OPT &  \bloom & T0 & mT0 & OPT \\
            \midrule
\multicolumn{5}{l}{WMT 2014} \\
\midrule
en$\rightarrow$fr & -0.985 & -0.700 & 0.453 & -0.919 &  0.085 & -1.035 & -0.015 & -0.165 \\
fr$\rightarrow$en & -0.675 & 0.337 & 0.567 & -0.493 &  0.448 & -0.087 & 0.250 & 0.039 \\
en$\rightarrow$hi & -0.482 & -1.819 & 0.484 & -1.525 &  0.288 & -1.733 & 0.026 & -1.460 \\
hi$\rightarrow$en & -0.387 & -1.346 & 0.514 & -1.200 &  0.378 & -1.624 & -0.019 & -1.290 \\
\midrule
\multicolumn{5}{l}{DiaBLa} \\
\midrule
en$\rightarrow$fr & -1.573 & -0.528 & 0.380 & -1.762 &  0.342 & -0.585 & -0.018 & 0.123 \\
fr$\rightarrow$en & -1.581 & 0.228 & 0.534 & -1.507 & 0.614 & -0.032 & 0.365 & 0.389 \\
\midrule
\multicolumn{5}{l}{Flores-101} \\
\midrule
en$\rightarrow$fr & -1.469 & -0.682 & 0.797 & -1.438 & 0.602 & -0.983 & 0.605 & 0.130 \\
fr$\rightarrow$en & -1.143 & 0.499 & 0.833 & -1.008 &  0.687 & -0.081 & 0.706 & 0.404 \\
en$\rightarrow$hi & -0.972 & -1.848 & 1.025 & -1.699 & 0.454 & -1.795 & 0.718 & -1.622 \\
hi$\rightarrow$en & -0.339 & -1.391 & 0.797 & -1.493 &  0.538 & -1.264 & 0.667 & -1.263 \\
\bottomrule
\end{tabular}}
\caption{\label{tab:xglm-main-orig-comet}
Original predictions}
\end{subtable}
\begin{subtable}[h]{0.48\textwidth}
            \centering\small
            \resizebox{\linewidth}{!}{
            \begin{tabular}{lrrrrrrrr}
            \toprule
            & \multicolumn{4}{c}{0-shot} & \multicolumn{4}{c}{1-shot} \\
            & \bloom & T0 & mT0 & OPT &  \bloom & T0 & mT0 & OPT \\
            \midrule
\multicolumn{5}{l}{WMT 2014} \\
\midrule
en$\rightarrow$fr & 0.434 & -0.700 & 0.452 & 0.034 & 0.424 & -1.035 & -0.017 & -0.000 \\
fr$\rightarrow$en & 0.604 & 0.336 & 0.566 & 0.534 & 0.532 & -0.090 & 0.247 & 0.449 \\
en$\rightarrow$hi & 0.053 & -1.819 & 0.483 & -1.491 & 0.448 & -1.733 & 0.026 & -1.460 \\
hi$\rightarrow$en & 0.445 & -1.346 & 0.511 & -1.113 & 0.386 & -1.624 & -0.022 & -1.274 \\
\midrule
\multicolumn{5}{l}{DiaBLa} \\
\midrule
en$\rightarrow$fr & 0.433 & -0.528 & 0.380 & -0.002 & 0.634 & -0.585 & -0.023 & 0.192 \\
fr$\rightarrow$en & 0.567 & 0.228 & 0.534 & 0.554 & 0.758 & -0.039 & 0.356 & 0.639 \\
\midrule
\multicolumn{5}{l}{Flores-101} \\
\midrule
en$\rightarrow$fr & 0.182 & -0.683 & 0.793 & 0.027 & 0.622 & -0.984 & 0.601 & 0.180 \\
fr$\rightarrow$en & 0.697 & 0.499 & 0.831 & 0.689 & 0.690 & -0.086 & 0.702 & 0.594 \\
en$\rightarrow$hi & -0.608 & -1.849 & 1.025 & -1.638 & 0.461 & -1.795 & 0.718 & -1.622 \\
hi$\rightarrow$en & 0.509 & -1.391 & 0.797 & -1.166 & 0.538 & -1.264 & 0.666 & -1.251 \\
\bottomrule
\end{tabular}}
\caption{\label{tab:xglm-main-truncated-comet}
Truncated predictions}
\end{subtable}
\caption{\label{tab:xglm-main-comparison-comet}Comparison of COMET scores across the three datasets using the \texttt{xglm-source+target} prompt.}
\end{table}

\section{Wrong language prediction and over-generation}\label{app:langid_overgen}

As described in Section~\ref{sec:dataset-comparison}, one problem identified with \bloom, particularly for 0-shot translation, is generating in the wrong language. Tables~\ref{tab:langid_wmt14_enfr} and \ref{tab:langid_wmt14_enhi} give the full analysis including raw figures for language identification for WMT14 fr$\leftrightarrow$en and hi$\leftrightarrow$en translation directions. For 0-5 few-shot examples, we indicate the number of truncated outputs identified as being from each language (indicated by the rows), the correct language (the target) being indicated in green, and the source language (therefore incorrect) being indicated in red. We also provide the average length difference ($\Delta$) between \bloom's outputs and the reference translations (negative numbers indicate that the prediction is longer than the reference).

For 0-shot translation, a significant number of examples are classed as being in the source language for en$\rightarrow$fr, and even more so for en$\rightarrow$hi (almost one fifth of the outputs are in the wrong language). As we increase the number of few-shot examples used, both of these problems are significantly reduced, and almost disappear for all language pairs and directions with 5 examples.

\begin{table}
  \begin{subtable}[h]{0.5\textwidth}
    \centering\small
    \resizebox{\linewidth}{!}{%
      \begin{tabular}{lrrrrrrrr}
        \toprule
        {} & \multicolumn{2}{c}{0-shot} & \multicolumn{2}{c}{1-shot} & \multicolumn{2}{c}{2-shot} & \multicolumn{2}{c}{5-shot} \\
        {} &  N &  $\Delta$ &  N &  $\Delta$ &  N &  $\Delta$ &  N & $\Delta$ \\
        \midrule
        cs &    1 & 408 &    - &   - &    - &   - &    - &  - \\
        de &    1 &   3 &    2 & 146 &    2 & -12.5 &    1 &  2 \\
        \rowcolor{OrangeRed!60}\textbf{en} &  181 &  16 &   32 &  57 &   10 &  73.8 &    8 & 92.2 \\
        es &    1 &  12 &    3 &  89.3 &    - &   - &    - &  - \\
        \rowcolor{SpringGreen!60}\textbf{fr} & 2814 &   7.9 & 2959 &   2.1 & 2989 &   1.5 & 2992 &  1.6 \\
        ht &    1 &  57 &    1 &  89 &    - &   - &    - &  - \\
        it &    2 &   4.5 &    3 &  13.3 &    - &   - &    - &  - \\
        nl &    1 & 131 &    - &   - &    - &   - &    - &  - \\
        pt &    1 & 146 &    - &   - &    - &   - &    - &  - \\
        ms &    - &   - &    1 &  28 &    - &   - &    - &  - \\
        ru &    - &   - &    1 &  16 &    - &   - &    - &  - \\
        zh &    - &   - &    1 &  10 &    - &   - &    - &  - \\
        ca &    - &   - &    - &   - &    1 & 198 &    1 & 18 \\
        uk &    - &   - &    - &   - &    1 &   3 &    1 &  3 \\
        \bottomrule
      \end{tabular}}
      \caption{\label{fig:langid-en2fr}en$\rightarrow$fr}
    \end{subtable}
    
    \begin{subtable}[h]{0.5\textwidth}
      \centering\small
      \resizebox{\linewidth}{!}{%
        \begin{tabular}{lrrrrrrrr}
          \toprule
          {} & \multicolumn{2}{l}{0-shot} & \multicolumn{2}{l}{1-shot} & \multicolumn{2}{l}{2-shot} & \multicolumn{2}{l}{5-shot} \\
          {} &  N &  $\Delta$ &  N & $\Delta$ &  N & $\Delta$ &  N & $\Delta$ \\
          \midrule
          \rowcolor{SpringGreen!60}\textbf{en} & 2954 &   1 & 2979 &  0.8 & 2988 &  1 & 2987 &  1.3 \\
          \rowcolor{OrangeRed!60}\textbf{fr} &   47 & -23.4 &   22 & -1.4 &   13 &  1.3 &   13 & -2.2 \\
          it &    1 &   3 &    - &  - &    2 &  6 &    3 &  5.3 \\
          tr &    1 &  -1 &    1 & -1 &    - &  - &    - &  - \\
          es &    - &   - &    1 &  1 &    - &  - &    - &  - \\
          \bottomrule
        \end{tabular}}
      \caption{\label{tab:langid-fr2en}fr$\rightarrow$en}
    \end{subtable}
  \caption{\label{tab:langid_wmt14_enfr} Raw figures for language identification and length differences of outputs compared to the reference translation for WMT2014 en$\rightarrow$fr using the \texttt{xglm-source+target} prompt. For 0-5 few-shot examples, N is the number of sentences identified as being in each language (the target language's row (correct) is indicated in green and the source language's row (one of the many incorrect options) in red) and $\Delta$ is the length difference in number of characters (N.B.~it is negative when the prediction is longer than the reference).}
\end{table}

\begin{table}
  \begin{subtable}[h]{0.5\textwidth}
    \centering\small
    \resizebox{\linewidth}{!}{%
      \begin{tabular}{lrrrrrrrr}
        \toprule
        {} & \multicolumn{2}{c}{0-shot} & \multicolumn{2}{c}{1-shot} & \multicolumn{2}{c}{2-shot} & \multicolumn{2}{c}{5-shot} \\
        {} &  N &   $\Delta$ &  N &  $\Delta$ &  N &  $\Delta$ &  N &  $\Delta$ \\
        \midrule
        ceb &    1 & -150 &    - &   - &    - &   - &    - &   - \\
        \rowcolor{OrangeRed!60}\textbf{en}  &  476 &   10.5 &   48 &  12.4 &   71 &  13.9 &   26 &  18.8 \\
        eo  &    1 & -134 &    - &   - &    - &   - &    - &   - \\
        fi  &    1 &   19 &    - &   - &    - &   - &    - &   - \\
        fr  &    2 &   94.5 &    - &   - &    - &   - &    - &   - \\
        gom &    2 &    6.5 &    1 &   4 &    - &   - &    1 &   0 \\
        \rowcolor{SpringGreen!60}\textbf{hi}  & 1998 &    9.3 & 2431 &   6 & 2403 &   5.5 & 2457 &   5.5 \\
        hsb &    1 &   98 &    - &   - &    - &   - &    - &   - \\
        ht  &    2 &  147 &    6 & 257.5 &   11 & 135.3 &    1 & 158 \\
        hu  &    1 &   71 &    - &   - &    - &   - &    - &   - \\
        lv  &    3 &   63.3 &    - &   - &    - &   - &    - &   - \\
        mr  &    5 &   64.4 &   11 &  14.6 &   17 &  11.7 &   19 &   6 \\
        ne  &    5 &    7.6 &    9 &  28.2 &    4 &  16.8 &    3 &   8.3 \\
        nl  &    2 &  -13.5 &    - &   - &    - &   - &    - &   - \\
        pt  &    1 &   24 &    - &   - &    - &   - &    - &   - \\
        sa  &    1 &  -25 &    - &   - &    - &   - &    - &   - \\
        sw  &    1 &   12 &    - &   - &    - &   - &    - &   - \\
        tl  &    1 &   24 &    - &   - &    - &   - &    - &   - \\
        war &    3 &    3 &    - &   - &    - &   - &    - &   - \\
        vec &    - &    - &    1 & -38 &    - &   - &    - &   - \\
        new &    - &    - &    - &   - &    1 &  25 &    - &   - \\
        \bottomrule
      \end{tabular}}%
      \caption{\label{fig:langid-en2hi}en$\rightarrow$hi}
  \end{subtable}
  \begin{subtable}[h]{0.5\textwidth}
    \centering\small
    \resizebox{\linewidth}{!}{
      \begin{tabular}{lrrrrrrrr}
        \toprule
        {} & \multicolumn{2}{c}{0-shot} & \multicolumn{2}{c}{1-shot} & \multicolumn{2}{c}{2-shot} & \multicolumn{2}{c}{5-shot} \\
        {} &  N &  $\Delta$ &  N &  $\Delta$ &  N &  $\Delta$ &  N &  $\Delta$ \\
        \midrule
        \rowcolor{SpringGreen!60}\textbf{en}  & 2469 &   4 & 2499 &   5.1 & 2503 &   3.8 & 2498 &   3 \\
        fr  &    1 & 151 &    1 &  -5 &    - &   - &    1 &   8 \\
        \rowcolor{OrangeRed!60}\textbf{hi}  &   29 &   3.3 &    2 &   0 &    - &   - &    - &   - \\
        ht  &    6 & 199.8 &    - &   - &    - &   - &    - &   - \\
        it  &    1 & 139 &    - &   - &    1 & -18 &    3 &   4.3 \\
        nl  &    1 &   9 &    - &   - &    - &   - &    2 &  -3 \\
        id  &    - &   - &    1 &  -6 &    - &   - &    - &   - \\
        nds &    - &   - &    1 &  16 &    - &   - &    - &   - \\
        pl  &    - &   - &    1 & -14 &    - &   - &    - &   - \\
        tr  &    - &   - &    1 & -15 &    - &   - &    - &   - \\
        war &    - &   - &    1 & 344 &    - &   - &    - &   - \\
        de  &    - &   - &    - &   - &    1 & -15 &    1 & 188 \\
        es  &    - &   - &    - &   - &    1 &   2 &    - &   - \\
        la  &    - &   - &    - &   - &    1 &  17 &    - &   - \\
        fi  &    - &   - &    - &   - &    - &   - &    1 &  -1 \\
        pt  &    - &   - &    - &   - &    - &   - &    1 &   1 \\
        \bottomrule
      \end{tabular}}
    \caption{\label{fig:langid-en2hi-orig}hi$\rightarrow$en}
  \end{subtable}
  \caption{\label{tab:langid_wmt14_enhi} Raw figures for language identification and length differences of outputs compared to the reference translation for WMT2014 en$\rightarrow$hi using the \texttt{xglm-source+target} prompt. For 0-5 few-shot examples, N is the number of sentences identified as being in each language (the target language's row (correct) is indicated in green and the source language's row (one of the many incorrect options) in red) and $\Delta$ is the length difference in number of characters (N.B.~it is negative when the prediction is longer than the reference).}
\end{table}

\section{Analysis per model \label{sec:permodel-appendix}}

In this section, we complete the results of Section~\ref{sec:model-size} with Tables~\ref{tab:per-model-analysis-big-bloom-en-fr-trunc} and \ref{tab:per-model-analysis-big-bloom-en-hi-trunc}, respectively for French$\leftrightarrow$English and Hindi$\leftrightarrow$English, reporting results without truncation. As expected, the systems are ranked according to their size. For French--English we see that decent performance can already be obtained with the second largest model \bloom-7b1, using 1-shot. Using this model, or even a model half this size can provide good indication of the performance of prompts, and be reliably used as test beds. We obtain less satisfactory results with English$\leftrightarrow$Hindi, even with the large \bloom; for this language pair, we even observe a large variation across prompts (looking at the range of scores) in the 1-shot setting for all models.

\begin{table*}[!ht]
\centering\small
\begin{tabular}{l*{4}{r@{ }c}}
  \toprule
  &      \multicolumn{4}{c}{0-shot} & 
               \multicolumn{4}{c}{1-shot}  \\
  Model / Direction & \multicolumn{2}{c}{en$\rightarrow$fr } & \multicolumn{2}{c}{fr$\rightarrow$en } & \multicolumn{2}{c}{en$\rightarrow$fr } & \multicolumn{2}{c}{fr$\rightarrow$en } \\
\midrule
  \bloom      &       \textbf{11.2} & {\scriptsize 3.0 -- 22.0} &    \textbf{15.4} &  {\scriptsize 10.3 -- 26.8} &   \textbf{32.6} & {\scriptsize 27.8 -- 36.4} & \textbf{34.9} &  {\scriptsize 33.1 -- 36.6} \\
  \bloom-7b1  &      6.5 & {\scriptsize 1.5 -- 12.1} &     12.8 &  {\scriptsize 4.8 -- 25.1}   &    25.9 & {\scriptsize 20.8 -- 29.9} & 29.1 &  {\scriptsize 25.4 -- 32.5} \\
  \bloom-3b   &      3.6 & {\scriptsize 1.2 -- 9.6} &       10.6 &  {\scriptsize 2.8 -- 19.3}   &    21.6 & {\scriptsize 16.7 -- 26.8} & 25.7 &  {\scriptsize 18.6 -- 29.6} \\
  \bloom-1b1  &     1.7 & {\scriptsize 0.5-- 3.9}   &        7.1 &  {\scriptsize 0.7 -- 11.4}    &    10.1 &  {\scriptsize 6.3 -- 13.2}  & 16.1 &  {\scriptsize 12.2 -- 19.9} \\
  \bloom-560m &    0.6 & {\scriptsize  0.4 -- 0.9}  &      3.7  &  {\scriptsize 1.4 -- 5.4}     &      3.6  &  {\scriptsize 2.2 -- 4.4}   &   8.6 &  {\scriptsize 5.8 -- 12.1} \\
  \bottomrule
\end{tabular}
 \caption{\label{tab:per-model-analysis-big-bloom-en-fr-trunc}Average, min and max BLEU scores per model of increasing size, for WMT14 en$\leftrightarrow$fr (original outputs). Best average result per setting in bold.}
\end{table*}

\begin{table*}[!ht]
\centering\small
\begin{tabular}{l*{4}{r@{ }c}}
  \toprule
  &      \multicolumn{4}{c}{0-shot} & 
               \multicolumn{4}{c}{1-shot}  \\
  Model / Direction & \multicolumn{2}{c}{en$\rightarrow$hi } & \multicolumn{2}{c}{hi$\rightarrow$en } & \multicolumn{2}{c}{en$\rightarrow$hi } & \multicolumn{2}{c}{hi$\rightarrow$en } \\
  \midrule 
  \bloom      &   \textbf{2.1} & {\scriptsize 0.3 -- 6.8} &  \textbf{8.3}  & {\scriptsize 0.7 -- 13.0} &  \textbf{12.9} & {\scriptsize 6.5 -- 14.6} &  \textbf{19.8} & {\scriptsize 10.0 -- 25.8} \\
  \bloom-7b1  &    0.1 & {\scriptsize 0.1 -- 3.0} &           5.7 & {\scriptsize 0.3 -- 9.5} &      5.9   &  {\scriptsize 0.3 -- 10.4} &  12.4 & {\scriptsize 1.0 -- 17.5} \\
  \bloom-3b   &    0.2 & {\scriptsize 0.0 -- 0.5} &            3.6  & {\scriptsize 0.0 -- 7.0} &     4.9   &  {\scriptsize 0.2 -- 7.2}   &   8.9 & {\scriptsize 0.1 -- 13.5} \\
  \bloom-1b1  &    0.1 & {\scriptsize 0.0 -- 0.1} &           1.5  & {\scriptsize 0.0 -- 4.5} &     1.4   &  {\scriptsize 0.1 -- 3.1}   &   4.6 & {\scriptsize 0.00 -- 8.2} \\
  \bloom-560m &   0.1 & {\scriptsize 0.0 -- 0.1} &           0.8 & {\scriptsize 0.0 -- 1.7} &      0.2  &  {\scriptsize 0.0 -- 0.3}  &   1.5  & {\scriptsize 0.1 --  2.8} \\
  \bottomrule
\end{tabular}
 \caption{\label{tab:per-model-analysis-big-bloom-en-hi-trunc}Average, min and max BLEU scores per model of decreasing size, for WMT14 en$\leftrightarrow$hi (original outputs). Best average result per setting in bold.}
\end{table*}

\section{Analysis per prompt \label{sec:perprompt-appendix}}
In this section, we replicate the analysis of Section~\ref{sec:per-prompt} and report results per prompt with truncated outputs in Tables~\ref{tab:per-prompt-analysis-big-bloom-en-fr-trunc} and \ref{tab:per-prompt-analysis-big-bloom-en-hi-trunc}. The conclusions are overall consistent with what we report for non-truncated outputs in the main text. We note that after truncating the outputs, \texttt{xglm-source+target} yields very good results across the board, outperforming its closest contenders  \texttt{a\_good\_translation-source+target} and \texttt{version-target} in almost all configurations. However, the choice of the prompt seems to matter more (a)~in the zero-shot setting, (b)~when translating out of English. Conversely our more stable results are for fr--en, 1-shot.

\begin{table*}[!ht]
\centering\small
\begin{tabular}{l*{4}{r@{ }c}}
\toprule
   &      \multicolumn{4}{c}{en$\rightarrow$fr} & 
               \multicolumn{4}{c}{fr$\rightarrow$en}  \\
  Prompt / Few-shot \# & \multicolumn{2}{c}{0} & \multicolumn{2}{c}{1} & \multicolumn{2}{c}{0} & \multicolumn{2}{c}{1} \\
\midrule 
\texttt{a\_good\_translation-source+target}  &  8.5 &  {\scriptsize 0.7--17.0} &   19.1 & {\scriptsize 4.32--37.12} &       16.4 &{\scriptsize  7.5--22.2} &         26.0 & {\scriptsize 12.0--37.0} \\
\texttt{a\_good\_translation-target} &   4.6 & {\scriptsize 0.6--13.9} &    20.9 & {\scriptsize 3.4--36.8} & 21.7 & {\scriptsize 6.6--35.2} &                26.31 & {\scriptsize 12.5--36.9} \\
\texttt{gpt3-target}  &             4.0 & {\scriptsize 0.7--14.0} &                 18.7 & {\scriptsize 3.0--36.4} &     8.3 & {\scriptsize 1.3--25.7} &        21.6 &  {\scriptsize 7.2--37.2} \\
\texttt{translate\_as-target} &     6.4 & {\scriptsize 0.6--10.1} &           18.1 & {\scriptsize 3.5--33.1} &    11.5 & {\scriptsize 2.3--20.4} &               22.9 &  {\scriptsize 8.2--35.7} \\
\texttt{version-target}      &      9.7 & {\scriptsize 0.7--30.3} &            21.9 & {\scriptsize 4.4--36.7} &     22.2 & {\scriptsize 4.7--35.2} &           25.3 &  {\scriptsize 8.0--37.2} \\
\texttt{xglm-source+target}  &      \textbf{17.2}& {\scriptsize 1.33--32.2} &   \textbf{23.2} & {\scriptsize 5.0--36.3} &  \textbf{25.6} & {\scriptsize 8.3--37.2} & \textbf{26.7} & {\scriptsize 11.1--38.2} \\
\texttt{xglm-target}         &      2.5 & {\scriptsize 1.1--4.6} &  20.1 & {\scriptsize 6.8--33.1} &     11.0 & {\scriptsize 4.5--17.6} &   23.1 & {\scriptsize 10.4--36.4} \\
\bottomrule
\end{tabular}
 \caption{\label{tab:per-prompt-analysis-big-bloom-en-fr-trunc}Average, min and max BLEU scores per prompt for WMT14 en$\leftrightarrow$fr (truncated outputs). Best average result per setting in bold.}
\
\end{table*}

\begin{table*}[!ht]
\centering\small
\begin{tabular}{l*{4}{r@{ }c}}
\toprule
   &      \multicolumn{4}{c}{en$\rightarrow$hi} & 
               \multicolumn{4}{c}{hi$\rightarrow$en}  \\
  Prompt / Few-shot \# & \multicolumn{2}{c}{0} & \multicolumn{2}{c}{1} & \multicolumn{2}{c}{0} & \multicolumn{2}{c}{1} \\\midrule
\texttt{a\_good\_translation-source+target} &  1.2 &{\scriptsize  0.1--3.3} &    5.8 & {\scriptsize 0.3--14.5} & 6.2 & {\scriptsize 1.0--12.7}   &                   13.0 & {\scriptsize 2.6--24.4} \\
\texttt{a\_good\_translation-target}     &   0.4 & {\scriptsize 0.1--1.3} &         5.5 & {\scriptsize 0.3--14.1} &     10.8 & {\scriptsize 1.1--25.4}  &            13.2 & {\scriptsize 2.7--24.7} \\
\texttt{gpt3-target}                  &     0.0 & {\scriptsize 0.0--0.1} &         1.6 & {\scriptsize 0.0--7.6}  &           0.0 & {\scriptsize 0.0--0.0}    &              2.5 & {\scriptsize 0.0--11.4} \\
\texttt{version-target}              &     1.0 & {\scriptsize 0.1--3.0} &  5.5 & {\scriptsize 0.2--13.9} & \textbf{11.3} & {\scriptsize 2.4--21.4}  &                    \textbf{13.5} & {\scriptsize 2.7--25.7} \\
\texttt{xglm-source+target}      &     \textbf{3.9} & {\scriptsize 0.1--12.1} &  \textbf{7.3} & {\scriptsize 0.2--15.8} &  8.8 & {\scriptsize 0.9--24.3}   &          12.4 & {\scriptsize 1.2--25.0} \\
\texttt{xglm-target}                 &    0.3 & {\scriptsize 0.0--1.0} &    5.1 & {\scriptsize 0.0--14.5} &   2.1 & {\scriptsize 0.3--5.8}     &               6.5 & {\scriptsize 0.1--13.0} \\
\bottomrule
\end{tabular}
 \caption{\label{tab:per-prompt-analysis-big-bloom-en-hi-trunc}Average, min and max BLEU scores per prompt for WMT14 en$\leftrightarrow$hi (truncated outputs). Best average result per setting in bold.}
\end{table*}

\section{Translation divergences in Flores 101}\label{sec:spanish-in-flores}

\begin{table*}[!ht]
  \centering\small
    \begin{tabular}{lp{0.9\textwidth}}
  \toprule
    en      & They are cooler than the surrounding surface in the day and warmer at night.    \\
    fr$\rightarrow$en & ``They are cooler than the surrounding surface during the day and warmer at night ''.    \\
    es$\rightarrow$en & During the day, its temperature is lower than that of the surrounding surface, and at night, higher.\\ 
    \midrule

    en       & ``This is not going to be goodbye. This is the closing of one chapter and the opening of a new one.''    \\
    fr$\rightarrow$en  & ``It’s not goodbye. It’s a page that is turning, and another that is opening.''    \\
    es$\rightarrow$en & "This will not be a farewell; it is just the end of one chapter and the beginning of another".\\
    \midrule

    en      &  ``We now have 4-month-old mice that are non-diabetic that used to be diabetic,'' he added.    \\
    fr$\rightarrow$en  & "We now have mice that are four months old and are not diabetic, whereas they were before", he added.   \\
    es$\rightarrow$en & ``Currently, we have mice that are four months old and used to be diabetic, but they are no longer diabetic'', he added. \\ 
    \midrule
    en      & ``We will endeavour to cut carbon dioxide emissions per unit of GDP by a notable margin by 2020 from the 2005 level,'' Hu said.   \\
    fr$\rightarrow$en  & ``We will strive to significantly reduce carbon dioxide emissions per unit of GDP by 2020 compared to the 2005 level,'' said Mr. Hu.    \\
    es$\rightarrow$en &  Hu said, ``We will work hard to reduce the level of carbon dioxide emitted per unit of GDP by 2020, so that the difference is significant compared to 2005.'' \\
    \midrule

    en      & Scientists say this animal's plumage was chestnut-brown on top with a pale or carotenoid-colored underside.    \\
    fr$\rightarrow$en  & Scientists say that the plumage of this animal was chestnut brown on top and pale or carotenoid on the underside. \\    
    es$\rightarrow$en & According to the experts, this animal has a brown plumage on the upper part and a pale or carotenoid color on the lower part. \\ 
    \midrule

    en      & 34 per cent of those in the poll share this view, wanting Queen Elizabeth II to be Australia's last monarch. \\
    fr$\rightarrow$en  & 34 \% of the people surveyed share this view, and want Queen Elizabeth II to be the last monarch to rule Australia.    \\
    es-en & One third of the respondents share this view and want the last queen to be Queen Elizabeth II.\\
    \bottomrule
  \end{tabular}
  \caption{\label{tab:spanish-french}A random subset of Flores-101 examples translated using \bloom into English from French and Spanish (N.B.~English was the original language of the sentences). Each block of three sentences contains the original English and the automatic French$\rightarrow$English and Spanish$\rightarrow$English translations.}
\end{table*}

A striking observation reported in the main text (Section~\ref{sec:per-lang-results}) is the difference between French and Spanish for the Flores-101 experiments. This is unexpected, as both languages are well represented in the training data. Yet, when translating from and into English the difference in spBLEU score is huge; and there is a clear gap with the other Romance languages as well. A related question is the poor translation between French and Spanish, not much better than for French$\rightarrow$Arabic. Looking at some sample outputs, this seems to be due to the peculiarities of the Spanish translations, which appear to be less literal than their French counterparts, but which yield equally good translations into English. This can be seen when we compare translations back into English for these languages (see a random subset in Table~\ref{tab:spanish-french}). The last example illustrates this very clearly: we see ``34 percent'' in both the original English and in the translation from French, while translation from Spanish starts with ``one third''.

\section{DiaBLa context-use examples}\label{app:diabla-examples}

Table~\ref{tab:diabla-examples} contains examples where the preceding context in 1-shot examples has a positive, negative or neutral influence on the current prediction, showing that the choice of the 1-shot example is important and is taken into account by the model. Some details of these experiments are found in the accompanying Section~\ref{sec:context} in the main text.
\begin{table*}[!ht]
    \centering\small
    \resizebox{\textwidth}{!}{
    \begin{tabular}{lp{10.9cm}p{2.5cm}p{3.2cm}}
    \toprule
    1-shot origin & Context & Reference & Prediction \\
    \midrule
    Prev. same & French: Pensez vous en permanence à \underline{la glace} qui se mange ? = English: Do you constantly think about the \underline{ice-cream} that's eaten? \#\#\# French: ou bien à \underline{une glace} pour se regarder ? = English:  & & *or to \underline{an ice-cream} to look at? \\
    Prev. opposite & English: Do you constantly think about the \underline{ice-cream} that's eaten? = French: Pensez vous en permanence à \underline{la glace} qui se mange ? \#\#\# French: ou bien à une glace pour se regarder ? = English: & Or a \underline{mirror} to look into? & or to a \underline{mirror} to look at? \\
    Random & French: N'empêche, on vit une époque folle, folle! = English: Still, what a crazy, crazy time we're living in! \#\#\# French: ou bien à \underline{une glace} pour se regarder ? = English: & &
    or to a \underline{mirror} to look at yourself? \\
    \midrule
    Prev. same & English: What kind of \underline{instrument} were you thinking of? = French: Tu penses à quelle sorte d'\underline{instrument}~? \#\#\# English: A wooden one I suppose... = French: &  & \underline{Un instrument} en bois, je suppose... \\
    Prev. opposite & French: Tu penses à quelle sorte d'\underline{instrument}~? = English: What kind of \underline{instrument} were you thinking of? \#\#\# English: A wooden one I suppose... = French: & \underline{Un instrument} en bois, je suppose. & \underline{Un instrument} en bois, je suppose... \\
    Random & French: Ils vont vous changer les idées après votre dure journée ! = English: They'll help you take your mind off things after your hard day! \#\#\# English: A wooden one I suppose... = French: & & \underline{Un} en bois, je suppose... \\
    \midrule
    Prev. same & English: He showed me how it works, but if I get stuck the \underline{girls} in here will always help me. = French: Il m'a montré comment cela marchait, mais si je n'y arrive pas, les \underline{filles} ici m'aideront sans problème. \#\#\# English: They are very kind. = French:	& & \underline{Elles} sont très \underline{gentilles}. \\
    Prev. opposite & French: Il m'a montré comment cela marchait, mais si je n'y arrive pas, les \underline{filles} ici m'aideront sans problème. = English: He showed me how it works, but if I get stuck the \underline{girls} in here will always help me. \#\#\# English: They are very kind. = French:	& \underline{Elles} sont très \underline{gentilles}. & \underline{Elles} sont très \underline{gentilles}. \\
    Random & English: I don't know about \underline{loans}. = French: Je ne sais pas pour les \underline{prêts}. \#\#\# English: They are very kind. = French: & & *\underline{Ils} sont très \underline{gentils}. \\
    \bottomrule
    \end{tabular}}
    \caption{\label{tab:diabla-examples}Ambiguous DiaBLa examples with different 1-shot contexts. Words that are relevant to the ambiguity are underlined, and incorrect translations are marked with an asterisk.}
\end{table*}

}
{}

\end{document}